# Multi-ellipses detection on images inspired by collective animal behavior

**Cuevas, E., González, M., Zaldívar, D., Pérez-Cisneros, M.**
Departamento de Ciencias Computacionales
Universidad de Guadalajara, CUCEI
Av. Revolución 1500, Guadalajara, Jal, México
{[1]erik.cuevas, mauricio.gonzalez }@cucei.udg.mx

## Abstract

This paper presents a novel and effective technique for extracting multiple ellipses from an image. The approach employs an evolutionary algorithm to mimic the way animals behave collectively assuming the overall detection process as a multi-modal optimization problem. In the algorithm, searcher agents emulate a group of animals that interact to each other using simple biological rules which are modeled as evolutionary operators. In turn, such operators are applied to each agent considering that the complete group has a memory to store optimal solutions (ellipses) seen so-far by applying a competition principle. The detector uses a combination of five edge points as parameters to determine ellipse candidates (possible solutions) while a matching function determines if such ellipse candidates are actually present in the image. Guided by the values of such matching functions, the set of encoded candidate ellipses are evolved through the evolutionary algorithm so that the best candidates can be fitted into the actual ellipses within the image. Just after the optimization process ends, an analysis over the embedded memory is executed in order to find the best obtained solution (the best ellipse) and significant local minima (remaining ellipses). Experimental results over several complex synthetic and natural images have validated the efficiency of the proposed technique regarding accuracy, speed and robustness.

## 1. Introduction

Ellipse is one of the most commonly occurring geometric shapes in real images. Even perfect circles in 3D space are projected into elliptical shapes in the image. Therefore, the extraction of ellipses from images is a key problem in computer vision and pattern recognition. Applications include, among others, human face detection [1], iris recognition [2], driving assistance [3], industrial applications [4] and traffic sign detection [5].

Ellipse detection in real images is an open research problem since long time ago. Several approaches have been proposed which traditionally fall under three categories: Symmetry-based, Hough transform-based (HT) and Random sampling.

In symmetry-based detection [5]-[7], the ellipse geometry is taken into account. The most common elements used in ellipse geometry are the ellipse center and axis. Using these elements and edges in the image, the ellipse parameters can be found. Ellipse detection in digital images is commonly solved through the Hough Transform [8]. It works by representing the geometric shape by its set of parameters, then accumulating bins in the quantized parameter space. Peaks in the bins provide the indication of where ellipses may be. Obviously, since the parameters are quantized into discrete bins, the intervals of the bins directly affect the accuracy of the results and the computational effort. Therefore, for fine quantization of the space, the algorithm returns more accurate results, while suffering from large memory loads and expensive computation. In order to overcome such a problem, some other researchers have proposed other ellipse detectors following the Hough transform principles by using random sampling. In random sampling-based approaches [9,10], a bin represents a candidate shape rather than a set of quantized parameters, as in the HT. However, like the HT, random sampling approaches go through an accumulation process for the bins. The bin with the highest score represents the best approximation of an actual ellipse in the target image. McLaughlin's work [11] shows that a random sampling-based approach produces improvements in accuracy and computational complexity, as well as a reduction in the number of false positives (non existent ellipses), when compared to the original HT and the number of its improved variants.

As an alternative to traditional techniques, the problem of ellipse detection has also been handled through optimization methods. In general, they have demonstrated to give better results than those based on the HT and random sampling with respect to accuracy, speed and robustness [12]. Such approaches have produced several robust ellipse detectors using different optimization algorithms such as Genetic algorithms (GA) [13,14] and Particle Swarm Optimization (PSO) [15]. In the detection, the employed optimization algorithms perform well in locating a single optimum (only one shape) but fail to provide multiple solutions. Since extracting multiple ellipses primitives falls into the category of multi-modal optimization, they need to be applied several times in order to extract all the primitives. This entails the removal of detected primitives from the image, one at a time, while iterating, until there are no more candidates left in the image which results in highly time consuming algorithms.

Multimodal optimization is used to locate all the optima within the searching space, rather than one and only one optimum, and has been extensively studied by many researchers [16]. Many algorithms based on a large variety of different techniques have been proposed in the literature. Among them, 'niches and species', and a fitness sharing method [17] have been introduced to overcome the weakness of traditional evolutionary algorithms for multimodal optimization. Here, a new optimization algorithm based on the collective animal behavior is proposed to solve multimodal problems, and then put into use in the application of multi-ellipse detection.

Many studies have been inspired by animal behavior phenomena in order to develop optimization techniques such as the Particle swarm optimization (PSO) algorithm which models the social behavior of bird flocking or fish schooling [18]. In recent years, there have been several attempts to apply the PSO to multi-modal function optimization problems [19,20]. However, the performance of such approaches presents several flaws when it is compared to the other multi-modal metaheuristic counterparts [21].

Recently, the concept of individual-organization [22, 23] has been widely used to understand collective behavior of animals. The central principle of individual-organization is that simple repeated interactions between individuals can produce complex behavioral patterns at group level [22, 24, 25]. Such inspiration comes from behavioral patterns seen in several animal groups, such as ant pheromone trail networks, aggregation of cockroaches and the migration of fish schools, which can be accurately described in terms of individuals following simple sets of rules [26]. Some examples of these rules [25, 27] include keeping current position (or location) for best individuals, local attraction or repulsion, random movements and competition for the space inside of a determined distance. On the other hand, new studies have also shown the existence of collective memory in animal groups [28]-[30]. The presence of such memory establishes that the previous history, of group structure, influences the collective behavior exhibited in future stages. Therefore, according to these new developments, it is possible to model complex collective behaviors by using simple individual rules and configuring a general memory.

This paper presents an algorithm for automatic detection of multiple ellipse shapes that considers the overall process as a multi-modal optimization problem. In the detection, the approach employs an evolutionary algorithm based on the way animals behave collectively. In such algorithm, searcher agents emulate a group of animals that interact to each other by simple biological rules which are modeled as evolution operators. Such operators are applied to each agent considering that the complete group has a memory which stores the optimal solutions seen so-far by applying a competition principle. The detector uses a combination of five edge points to determine ellipse candidates (possible solutions). A matching function determines if such ellipse candidates are actually present in the image. Guided by the values of such matching functions, the set of encoded candidate ellipses are evolved through the evolutionary algorithm so that the best candidates can be fitted into the actual ellipse within the image. Following the optimization process, the embedded memory is analyzed in order to find the best ellipse and significant local minima (remaining ellipses). Experimental results over several complex synthetic and natural images have validated the efficiency of the proposed technique regarding accuracy, speed and robustness.

The paper is organized as follows: Section 2 provides information regarding the evolutionary algorithm based on the way animals behave collectively. Section 3 depicts the implementation of the proposed ellipse detector. The complete multiple ellipse detection procedure is presented by Section 4. Experimental results for the proposed approach are stated in Section 5 and some relevant conclusions are discussed in Section 6.

## 2. Collective Animal Behavior Algorithm (CAB)

The CAB algorithm assumes the existence of a set of operations that resembles the interaction rules that model the collective animal behavior. In the approach, each solution within the search space represents an animal position. The "fitness value" refers to the animal dominance with respect to the group. The complete process mimics the collective animal behavior.

The approach in this paper implements a memory for storing best solutions (animal positions) mimicking the aforementioned biologic process. Such memory is divided into two different elements, one for maintaining the best locations at each generation ($\mathbf{M}_g$) and the other for storing the best historical positions during the complete evolutionary process ($\mathbf{M}_h$).

*2.1 Description of the CAB algorithm*

Following other metaheuristic approaches, the CAB algorithm is an iterative process that starts by initializing the population randomly (generated random solutions or animal positions). Then, the following four operations are applied until a termination criterion is met (i.e. the iteration number *NI*):

1. Keep the position of the best individuals.
2. Move from or to nearby neighbors (local attraction and repulsion).
3. Move randomly.
4. Compete for the space within a determined distance (update the memory).

*2.1.1 Initializing the population*

The algorithm begins by initializing a set $\mathbf{A}$ of $N_p$ animal positions ($\mathbf{A} = \{\mathbf{a}_1, \mathbf{a}_2, \ldots, \mathbf{a}_{N_p}\}$). Each animal position $\mathbf{a}_i$ is a *D*-dimensional vector containing parameter values to be optimized. Such values are randomly and uniformly distributed between the pre-specified lower initial parameter bound $a_j^{low}$ and the upper initial parameter bound $a_j^{high}$.

$$a_{j,i} = a_j^{low} + \text{rand}(0,1) \cdot (a_j^{high} - a_j^{low}); \qquad (1)$$
$$j = 1, 2, \ldots, D; \quad i = 1, 2, \ldots, N_p.$$

with *j* and *i* being the parameter and individual indexes respectively. Hence, $a_{j,i}$ is the *j*th parameter of the *i*th individual.

All the initial positions $\mathbf{A}$ are sorted according to the fitness function (dominance) to form a new individual set $\mathbf{X} = \{\mathbf{x}_1, \mathbf{x}_2, \ldots, \mathbf{x}_{N_p}\}$, so that we can choose the best *B* positions and store them in the memory $\mathbf{M}_g$ and $\mathbf{M}_h$. The fact that both memories share the same information is only allowed at this initial stage.

*2.1.2 Keep the position of the best individuals.*

Analogous to the biological metaphor, this behavioral rule, typical from animal groups, is implemented as an evolutionary operation in our approach. In this operation, the first $B$ elements ($\{\mathbf{a}_1, \mathbf{a}_2, \ldots, \mathbf{a}_B\}$), of the new animal position set $\mathbf{A}$, are generated. Such positions are computed by the values contained inside the historical memory $\mathbf{M}_h$, considering a slight random perturbation around them. This operation can be modeled as follows:

$$\mathbf{a}_l = \mathbf{m}_h^l + \mathbf{v} \qquad (2)$$

where $l \in \{1, 2, \ldots, B\}$ while $\mathbf{m}_h^l$ represents the $l$-element of the historical memory $\mathbf{M}_h$. $\mathbf{v}$ is a random vector with a small enough length.

*2.1.3 Move from or to nearby neighbors.*

From the biological inspiration, animals experiment a random local attraction or repulsion according to an internal motivation. Therefore, we have implemented new evolutionary operators that mimic such biological pattern. For this operation, a uniform random number $r_m$ is generated within the range [0,1]. If $r_m$ is less than a threshold $H$, a determined individual position is moved (attracted or repelled) considering the nearest best historical position within the group (i.e. the nearest position in $\mathbf{M}_h$); otherwise, it goes to the nearest best location within the group for the current generation (i.e. the nearest position in $\mathbf{M}_g$). Therefore such operation can be modeled as follows:

$$\mathbf{a}_i = \begin{cases} \mathbf{x}_i \pm r \cdot (\mathbf{m}_h^{nearest} - \mathbf{x}_i) & \text{with probability } H \\ \mathbf{x}_i \pm r \cdot (\mathbf{m}_g^{nearest} - \mathbf{x}_i) & \text{with probability } (1-H) \end{cases} \qquad (3)$$

where $i \in \{B+1, B+2, \ldots, N_p\}$, $\mathbf{m}_h^{nearest}$ and $\mathbf{m}_g^{nearest}$ represent the nearest elements of $\mathbf{M}_h$ and $\mathbf{M}_g$ to $\mathbf{x}_i$, while $r$ is a random number.

*2.1.4 Move randomly.*

Following the biological model, under some probability $P$, one animal randomly changes its position. Such behavioral rule is implemented considering the next expression:

$$\mathbf{a}_i = \begin{cases} \mathbf{r} & \text{with probability } P \\ \mathbf{x}_i & \text{with probability } (1-P) \end{cases} \qquad (4)$$

being $i \in \{B+1, B+2, \ldots, N_p\}$ and $\mathbf{r}$ a random vector defined in the search space. This operator is similar to re-initialize the particle in a random position, as it is done by Eq. (1).

*2.1.5. Compete for the space within of a determined distance (update the memory).*

Once the operations to keep the position of the best individuals, such as moving from or to nearby neighbors and moving randomly, have all been applied to the all $N_p$ animal positions, generating $N_p$ new positions, it is necessary to update the memory $\mathbf{M}_h$.

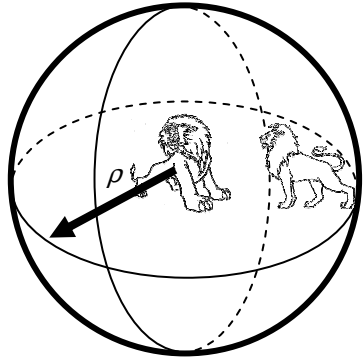

**Fig. 1.** Dominance concept as it is presented when two animals confront each other inside of a $\rho$ distance.

The concept of dominance is used to update the memory $M_h$. Animals that interact within a group maintain a minimum distance among them. Such distance $\rho$ depends on how aggressive an animal behaves [31,32]. Hence, when two animals confront each other inside such distance, the most dominant individual prevails meanwhile the other withdraws. Figure 1 depicts the process.

In the proposed algorithm, the historical memory $M_h$ is updated considering the following procedure:

1. The elements of $M_h$ and $M_g$ are merged into $M_U$ ($M_U = M_h \cup M_g$).
2. Each element $m_U^i$ of the memory $M_U$ is compared pair-wise to remaining memory elements ($\{m_U^1, m_U^2, \ldots, m_U^{2B-1}\}$). If the distance between both elements is less than $\rho$, the element getting a better performance in the fitness function prevails meanwhile the other is removed.
3. From the resulting elements of $M_U$ (from step 2), it is selected the *B* best value to build the new $M_h$.

Unsuitable values of $\rho$ yield a lower convergence rate, a longer computational time, a larger function evaluation number, the convergence to a local maximum or to an unreliable solution. The $\rho$ value is computed considering the following equation:

$$\rho = \frac{\prod_{j=1}^{D}(a_j^{high} - a_j^{low})}{10 \cdot D} \qquad (5)$$

where $a_j^{low}$ and $a_j^{high}$ represent the pre-specified lower and upper bound of the *j*-parameter respectively, in an *D*-dimensional space.

*2.1.6. Computational procedure*

The computational procedure for the proposed algorithm can be summarized as follows:

Step 1: Set the parameters $N_p$, *B*, *H*, *P* and *NI*.
Step 2: Generate randomly the position set $A = \{a_1, a_2, \ldots, a_{N_p}\}$ using Eq.1
Step 3: Sort **A** according to the objective function (dominance) to build $X = \{x_1, x_2, \ldots, x_{N_p}\}$.

Step 4:      Choose the first *B* positions of **X** and store them into the memory $\mathbf{M}_g$.

Step 5:      Update $\mathbf{M}_h$ according to Section 2.1.5. (during the first iteration: $\mathbf{M}_h = \mathbf{M}_g$).

Step 6:      Generate the first *B* positions of the new solution set **A** ($\{\mathbf{a}_1, \mathbf{a}_2, \ldots, \mathbf{a}_B\}$). Such positions correspond to the elements of $\mathbf{M}_h$ making a slight random perturbation around them.

$$\mathbf{a}_l = \mathbf{m}_h^l + \mathbf{v}$$ ; being **v** a random vector of a small enough length.

Step 7:      Generate the rest of the **A** elements using the attraction, repulsion and random movements.

         for $i = B+1 : N_p$
            if ($r_1 < P$) then
            *attraction and repulsion movement*
            { if ($r_2 < H$) then
$$\mathbf{a}_i = \mathbf{x}_i \pm r \cdot (\mathbf{m}_h^{nearest} - \mathbf{x}_i)$$
              else if
$$\mathbf{a}_i = \mathbf{x}_i \pm r \cdot (\mathbf{m}_g^{nearest} - \mathbf{x}_i)$$
            }
            else if
            *random movement*
            {
$$\mathbf{a}_i = \mathbf{r}$$
            }
        end for    where $r_1, r_2, r \in \text{rand}(0,1)$

Step 8:      If *NI* is completed, the process is finished; otherwise go back to step 3.
                The best value in $\mathbf{M}_h$ represents the global solution for the optimization problem.

**3. Ellipse detection using CAB**

*3.1. Data preprocessing*

In order to detect ellipse shapes, candidate images must be preprocessed first by the well-known Canny algorithm which yields a single-pixel edge-only image. Then, the $(x_i, y_i)$ coordinates for each edge pixel $p_i$ are stored inside the edge vector $P = \{p_1, p_2, \ldots, p_{N_p}\}$, with $N_p$ being the total number of edge pixels.

*3.2. Individual representation*

Each candidate solution *E* (ellipse candidate) uses five edge points. Under such representation, edge points are selected following a random positional index within the edge array *P*. This procedure will encode a candidate solution as the ellipse that passes through five points $p_i, p_j, p_k, p_l$ and $p_m$ ($E = \{p_i, p_j, p_k, p_l, p_m\}$). Thus, by substituting the coordinates of each point of *E* into Eq. 6, we gather a set of five simultaneous equations which are linear in the five unknown parameters $a', b', c', f'$ and $g'$.

$$a'x^2 + 2h'xy + b'y^2 + 2g'x + 2f'y + c' = 0 \qquad (6)$$

Then, solving the involved parameters and dividing by the constant $c'$, it yields:

$$ax^2 + 2hxy + by^2 + 2gx + 2fy + 1 = 0 \tag{7}$$

Considering the configuration of the edge points shown by Figure 2, the ellipse center $(x_0, y_0)$, the radius maximum ($r_{max}$), the radius minimum ($r_{min}$) and the ellipse orientation ($\theta$) can be calculated as follows:

$$x_0 = \frac{hf - bg}{C}, \tag{8}$$

$$y_0 = \frac{gh - af}{C}, \tag{9}$$

$$r_{max} = \sqrt{\frac{-2\Delta}{C(a+b-R)}}, \tag{10}$$

$$r_{min} = \sqrt{\frac{-2\Delta}{C(a+b+R)}}, \tag{11}$$

$$\theta = \frac{1}{2}\arctan\left(\frac{2h}{a-b}\right) \tag{12}$$

where

$$R^2 = (a-b)^2 + 4h^2 \text{ and } C = ab - h^2 \tag{13}$$

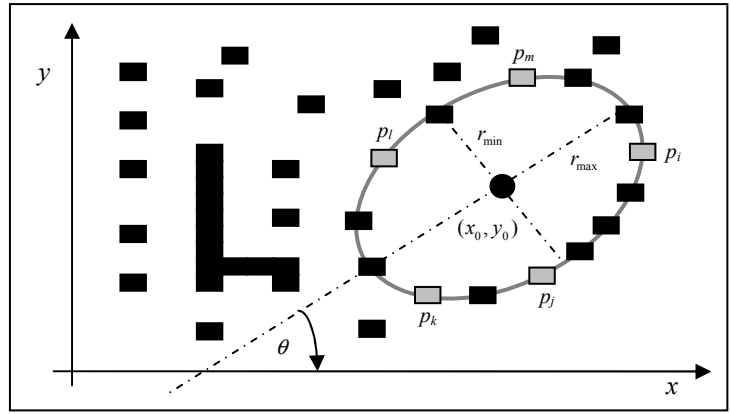

**Fig. 2.** Ellipse candidate (individual) built from the combination of points $p_i$, $p_j$, $p_k$, $p_l$ and $p_m$.

*3.3 Objective function*

Optimization refers to choosing the best element from one set of available alternatives. In the simplest case, it means to minimize an objective function or error by systematically choosing the values of variables from their valid ranges. In order to calculate the error produced by a candidate solution *E*, the ellipse coordinates are calculated as a virtual shape which, in turn, must also be validated, i.e. if it really exists in the edge image. The test set is represented by $S = \{s_1, s_2, \ldots, s_{N_s}\}$, where $N_s$ are the number of points over which the existence of an edge point, corresponding to *E*, should be tested.

The set $S$ is generated by the Midpoint Ellipse Algorithm (MEA) [33] which is a searching method that seeks required points for drawing an ellipse. Any point $(x, y)$ on the boundary of the ellipse with $a,h,b,g$ and $f$ satisfies the equation $f_{ellipse}(x,y) \cong r_x^2 x^2 + r_y^2 y^2 - r_x^2 r_y^2$. However, MEA avoids computing square-root calculations by comparing the pixel separation distances. A method for direct distance comparison is to test the halfway position between two pixels (sub-pixel distance) to determine if this midpoint is inside or outside the ellipse boundary. If the point is in the interior of the ellipse, the ellipse function is negative. Thus, if the point is outside the ellipse, the ellipse function is positive. Therefore, the error involved in locating pixel positions using the midpoint test is limited to one-half the pixel separation (sub-pixel precision). To summarize, the relative position of any point $(x, y)$ can be determined by checking the sign of the ellipse function:

$$f_{Circle}(x,y) \begin{cases} < 0 & \text{if } (x,y) \text{ is inside the ellipse boundary} \\ = 0 & \text{if } (x,y) \text{ is on the ellipse boundary} \\ > 0 & \text{if } (x,y) \text{ is outside the ellipse boundary} \end{cases} \qquad (14)$$

The ellipse-function test in Eq. 14 is applied to mid-positions between pixels nearby the ellipse path at each sampling step. Figure 3a and 4a are shows the midpoint between the two candidate pixels at sampling position. The ellipse is used to divide the quadrants into two regions the limit of the two regions is the point at which the curve has a slope of -1 as shown in Figure 4.

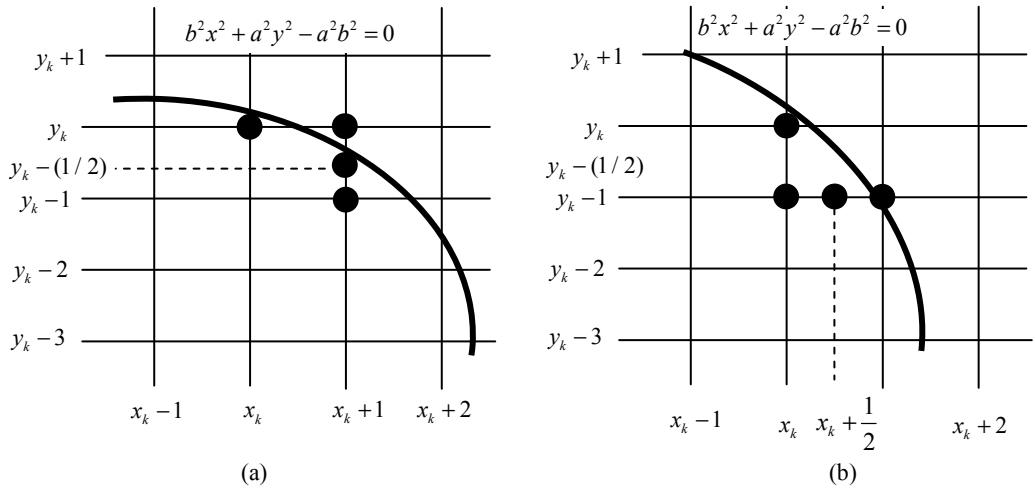

**Fig. 3.** (a) symmetry of the ellipse: an estimated one octant which belong to the first region where the slope is greater than -1, b) In this region the slope will be less than -1 to complete the octant and continue to calculate the same so the remaining octants.

In MEA the computation time is reduced by considering the symmetry of ellipses. Ellipses sections in adjacent octants within one quadrant are symmetric with respect to the dy/dy=-1 line dividing the two octants. These symmetry conditions are illustrated in Figure 4. The algorithm can be considered as the quickest providing a sub-pixel precision [34]. However, in order to protect the MEA operation, it is important to assure that points lying outside the image plane must not be considered in S.

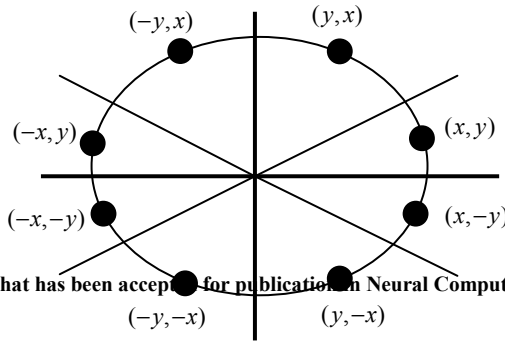

**Fig. 4.** Midpoint between candidate pixels at sampling position $x_k$ along an elliptical path

The objective function $J(E)$ represents the matching error produced between the pixels $S$ of the ellipse candidate $E$ and the pixels that actually exist in the edge image, yielding:

$$J(E) = 1 - \frac{\sum_{v=1}^{Ns} G(x_v, y_v)}{Ns} \quad (15)$$

where $G(x_i, y_i)$ is a function that verifies the pixel existence in $(x_v, y_v)$, with $(x_v, y_v) \in S$ and $N_s$ being the number of pixels lying on the perimeter corresponding to $E$ currently under testing. Hence, function $G(x_v, y_v)$ is defined as:

$$G(x_v, y_v) = \begin{cases} 1 & \text{if the pixel } (x_v, y_v) \text{ is an edge point} \\ 0 & \text{otherwise} \end{cases} \quad (16)$$

A value of $J(E)$ near to zero implies a better response from the "ellipsoid" operator. Figure 5 shows the procedure to evaluate a candidate action $E$ with its representation as a virtual shape $S$. Figure 5(a) shows the original edge map, while Figure 5(b) presents the virtual shape $S$ representing the individual $E = \{p_i, p_j, p_k, p_l, p_m\}$. In Figure 5(c), the virtual shape $S$ is compared to the original image, point by point, in order to find coincidences between virtual and edge points. The individual has been built from points $p_i$, $p_j$, $p_k$, $p_l$ and $p_m$ which are shown by Fig. 5(a). The virtual shape $S$, obtained by MEA, gathers 52 points ($N_s$ = 52) with only 35 of them existing in both images (shown as darker points in Fig. 5(c)) and yielding: $\sum_{v=1}^{Ns} G(x_v, y_v) = 35$, therefore $J(E)$=0.327.

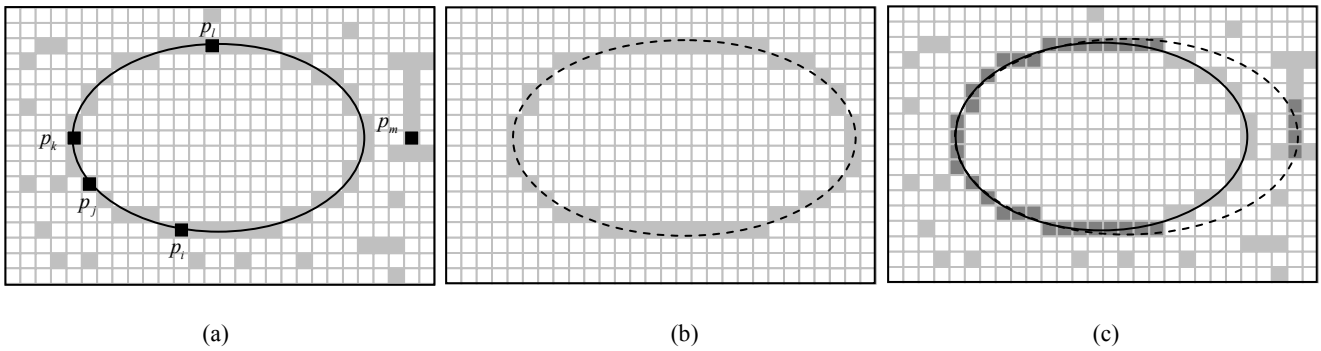

(a)          (b)          (c)

**Fig. 5.** Evaluation of a candidate solution $E$: the image in (a) shows the original image while (b) presents the generated virtual shape drawn from points $p_i$, $p_j$, $p_k$, $p_l$ and $p_m$. The image in (c) shows coincidences between both images which have been marked by darker pixels while the virtual shape is also depicted through a dashed line

*3.4 Implementation of CAB for ellipse detection*

The implementation of the proposed algorithm can be summarized in the following steps:

Step 1: Adjust the algorithm parameters $N_p$, $B$, $H$, $P$ and $NI$.

Step 2: Randomly generate a set of candidate ellipses (position of each animal) $\mathbf{E} = \{E_1, E_2, \ldots, E_{N_p}\}$ set using Eq.1

Step 3: Sort $\mathbf{E}$ according to the objective function (dominance) to build $\mathbf{X} = \{\mathbf{x}_1, \mathbf{x}_2, \ldots, \mathbf{x}_{N_p}\}$.

Step 4: Choose the first $B$ positions of $\mathbf{X}$ and store them into the memory $\mathbf{M}_g$.

Step 5: Update $\mathbf{M}_h$ according to Section 2.1.5 (during the first iteration: $\mathbf{M}_h = \mathbf{M}_g$).

Step 6: Generate the first $B$ positions of the new solution set $\mathbf{E}$ ($\{E_1, E_2, \ldots, E_B\}$). Such positions correspond to the elements of $\mathbf{M}_h$ making a slight random perturbation around them.

$$E_l = \mathbf{m}_h^l + \mathbf{v}$$ ; being $\mathbf{v}$ a random vector of a small enough length.

Step 7: Generate the rest of the $\mathbf{E}$ elements using the attraction, repulsion and random movements.
 for $i = B+1 : N_p$
  if ($r_1 < P$) then
   *attraction and repulsion movement*
   { if ($r_2 < H$) then
    $E_i = \mathbf{x}_i \pm r \cdot (\mathbf{m}_h^{nearest} - \mathbf{x}_i)$
    else if
    $E_i = \mathbf{x}_i \pm r \cdot (\mathbf{m}_g^{nearest} - \mathbf{x}_i)$
   }
   else if
   *random movement*
   {
    $E_i = \mathbf{r}$
   }
  end for where $r_1, r_2, r \in \text{rand}(0,1)$

Step 8: If $NI$ is completed, the process is finished; otherwise go back to step 3. The best values in $\mathbf{M}_h$ represents the best solutions (the best found ellipses).

## 4. The multiple ellipse detection procedure.

Most detectors simply apply a one-minimum optimization algorithm to detect multiple ellipses yielding only one ellipse at a time and repeating the same process several times as previously detected primitives are removed from the image. The algorithm iterates until no more candidates are left in the image.

On the other hand, the method at this paper is able to detect single or multiples ellipses through only one optimization step. The multi-detection procedure can be summarized as follows: guided by the values of a matching function, the whole group of encoded candidate ellipses is evolved through the set of evolutionary operators. The best ellipse candidate (global optimum) is considered to be the first detected ellipse over the edge-only image. An analysis of the incorporated historical memory $\mathbf{M}_h$ is thus executed in order to identify other local optima (other ellipses).

In order to find other possible ellipses contained in the image, the historical memory $\mathbf{M}_h$ is carefully examined. The approach aims to explore all elements, one at a time, assessing which of them represents an actual ellipse in the image. Since several elements can represent the same ellipse (i.e. ellipses slightly shifted or holding small deviations), a distinctiveness factor $D_{A,B}$ is required to measure the mismatch between two given ellipses (*A* and *B*). Such distinctiveness factor is defined as follows:

$$D_{A,B} = \left|x_0^A - x_0^B\right| + \left|y_0^A - y_0^B\right| + \left|r_{min}^A - r_{min}^B\right| + \left|r_{max}^A - r_{max}^B\right| + \left|\theta_A - \theta_B\right| \tag{17}$$

being $\left(x_0^A, y_0^A\right), r_{min}^A, r_{max}^A$ and $\theta_A$, the central coordinates, the lengths of the semi-axes and the orientation of the ellipse $E_A$, respectively. Likewise, $\left(x_0^B, y_0^B\right), r_{min}^B, r_{max}^B$ and $\theta_B$ represent the corresponding parameters of the ellipse $E_B$. One threshold value *Th* is also calculated to decide whether two ellipses must be considered different or not. *Th* is computed as:

$$Th = \frac{\left(\left|r_{max}^h - r_{max}^l\right| + \left|r_{min}^h - r_{min}^l\right|\right)/2}{s} \tag{18}$$

where $\left[r_{max}^l, r_{max}^h\right]$ and $\left[r_{min}^l, r_{min}^h\right]$ are the feasible semi-axes ranges and *s* is a sensitivity parameter. By using a high value for *s*, two very similar ellipses would be considered different while a smaller value for *s* would consider them as similar. In this work, after several experiments, the *s* value has been set to 2.

Thus, since the historical memory $\mathbf{M}_h$ $\left\{E_1^\mathbf{M}, E_2^\mathbf{M}, \ldots, E_B^\mathbf{M}\right\}$ groups the elements in descending order according to their fitness values, the first element $E_1^\mathbf{M}$, whose fitness value represents the best value $F_1^\mathbf{M}$, is assigned to the first ellipse. Then, the distinctiveness factor ($D_{E_1^\mathbf{M}, E_2^\mathbf{M}}$) over the next element $E_2^\mathbf{M}$ is evaluated with respect to the prior $E_1^\mathbf{M}$. If $D_{E_1^\mathbf{M}, E_2^\mathbf{M}} > Th$, then $E_2^\mathbf{M}$ is considered as a new ellipse otherwise the next element $E_3^\mathbf{M}$ is selected. This process is repeated until the fitness value $F_i^\mathbf{M}$ reaches a minimum threshold $F_{TH}$. According to such threshold, other values above $F_{TH}$ represent individuals (ellipses) that are considered as significant while other values lying below such boundary are considered as false ellipses and hence they are not contained in the image. After several experiments the value of $F_{TH}$ is set to $(F_1^\mathbf{M}/10)$.

## 5. Experimental results

Experimental tests have been developed in order to evaluate the performance of the ellipse detector. The experiments address the following tasks:

(1) Ellipse localization,
(2) Shape discrimination,
(3) Ellipse approximation: occluded ellipse and ellipsoidal detection.

Table 1 presents the parameters for the CAB algorithm at this work. They have been kept for all test images after being experimentally defined.

| $N_p$ | $H$ | $P$ | $B$ | $NI$ |
|---|---|---|---|---|
| 30 | 0.5 | 0.1 | 12 | 200 |

**Table 1.** CAB detector parameters

*5.1 Ellipse localization*

*5.1.1 Synthetic images*

The experimental setup includes the use of several synthetic images of 400x300 pixels. All images contain a different amount of ellipsoidal shapes and some have also been contaminated by added noise as to increase the complexity of the localization task. The algorithm is executed over 50 times for each test image, successfully identifying and marking all required ellipse in the image. The detection has proved to be robust to translation and scaling still offering a reasonably low execution time. Figure 6 shows the outcome after applying the algorithm to two images from the experimental set.

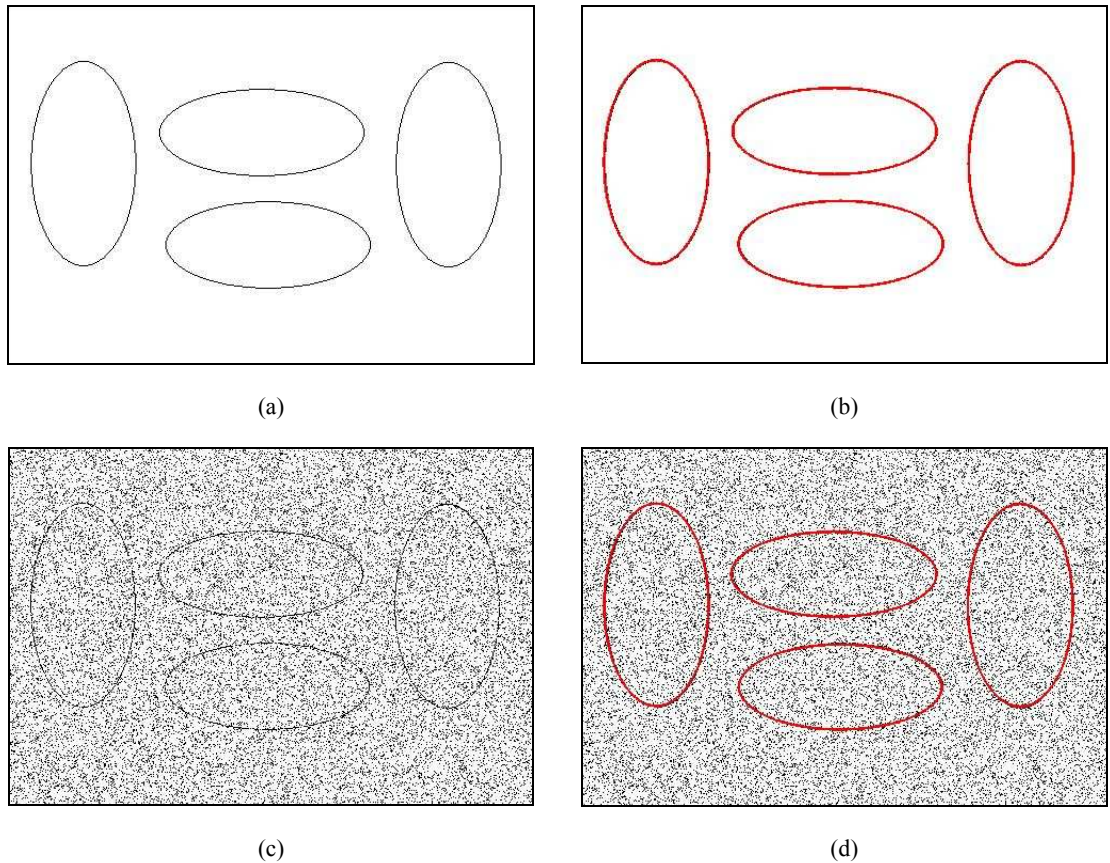

Fig. 6. Ellipse localization over synthetic images. The image (a) shows the original image while (b) presents the detected ellipses. The image in (c) shows a second image with salt & pepper noise and (d) shows the detected ellipses in red overlay.

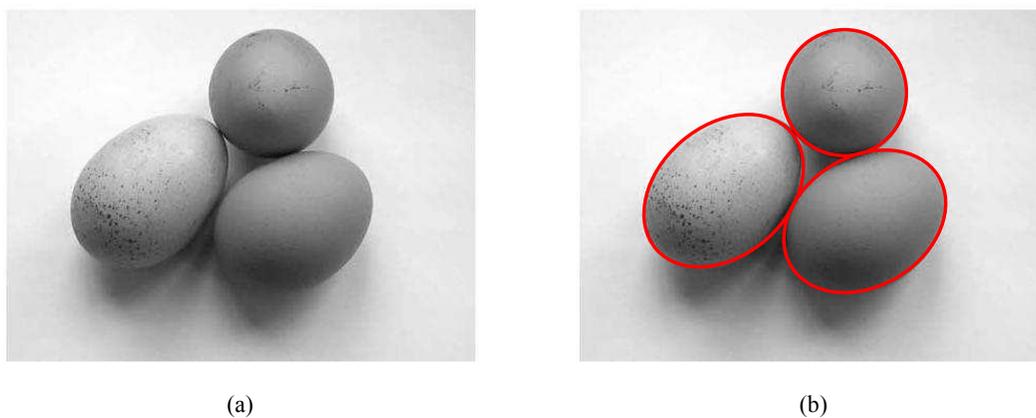

Fig. 7. Ellipse detection algorithm over natural images: the image in (a) shows the original image while (b) presents the detected ellipses in a red overlay.

*5.1.2 Natural images*

This experiment tests the ellipse detection on several real images which contain a different number of ellipse shapes. The image set has been captured by a digital camera in an 8-bit format. Each image is pre-processed by the Canny edge detection algorithm. Figure 7 shows the results after the algorithm CAB has been applied to one image from the experimental set.

*5.2 Shape discrimination tests*

This section discusses on the algorithm's ability to detect elliptical patterns despite some other different shapes being present in the image. Figure 8 shows four shapes in the image of 400x300 pixels. The image has been contaminated by local noise in order to increase the complexity on the localization task. Figure 9 repeats the experiment over a real-life image.

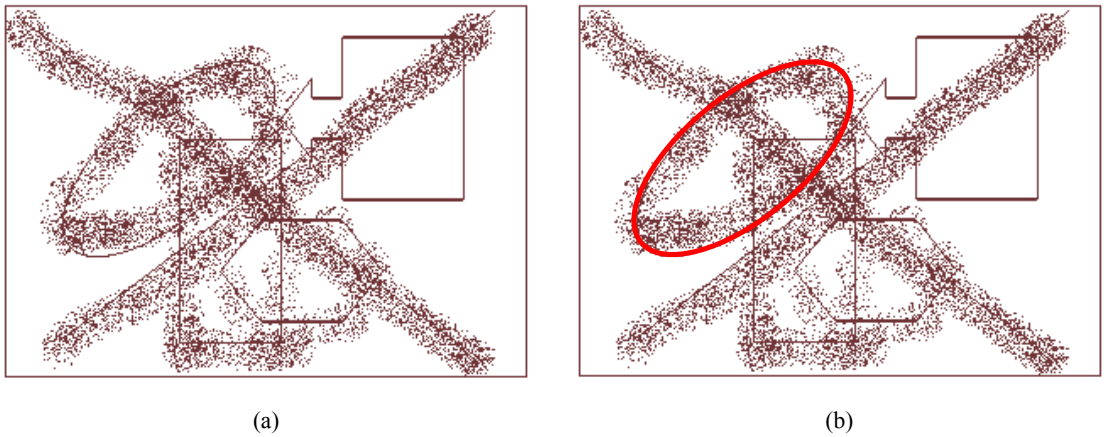

(a)            (b)

**Fig. 8.** Shape discrimination over synthetic images: (a) shows the original image (b) presents the detected ellipse as a red overlay.

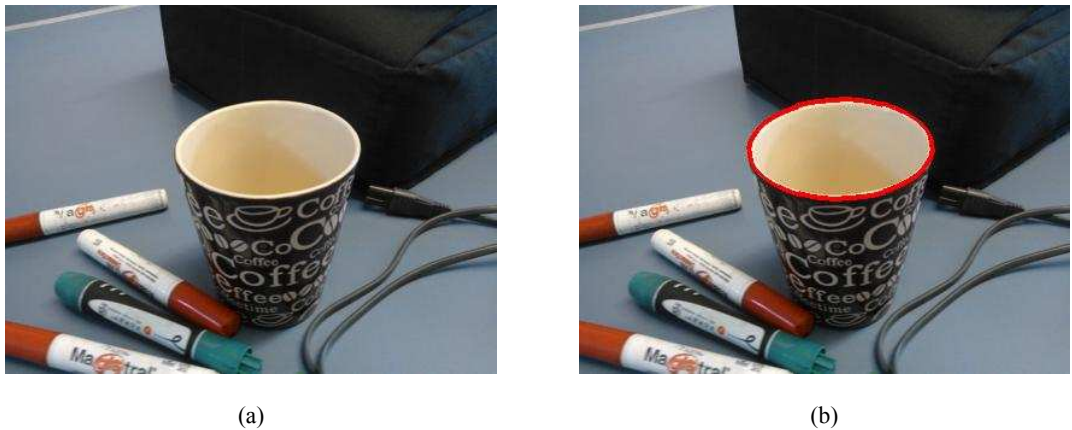

(a)            (b)

**Fig 9.** Shape discrimination in real-life images: (a) shows the original image and (b) presents the detected ellipse as an overlay.

*5.3 Ellipse approximation: occluded ellipse and ellipsoidal detection.*

The CAB detector algorithm is able to detect occluded or imperfect ellipses as well as partially defined shapes such as arc segments. The relevance of such functionality comes from the fact that imperfect ellipses are commonly found in typical computer vision applications. Since ellipse detection has been considered as an optimization problem, the CAB algorithm allows finding ellipses that may approach a given shape according to fitness values for each candidate. Figure 10(a) shows some examples of ellipse approximation. Likewise, the proposed algorithm is able to find ellipse parameters that better approach an arc or an occluded ellipse. Figure 10(b) and 10(c) show some examples of this functionality. A small value for $J(C)$, i.e., near zero, refers to an ellipse while a slightly bigger value accounts for an arc or an occluded circular shape. Such a fact does not represent any problem as ellipses can be shown following the obtained $J(C)$ values.

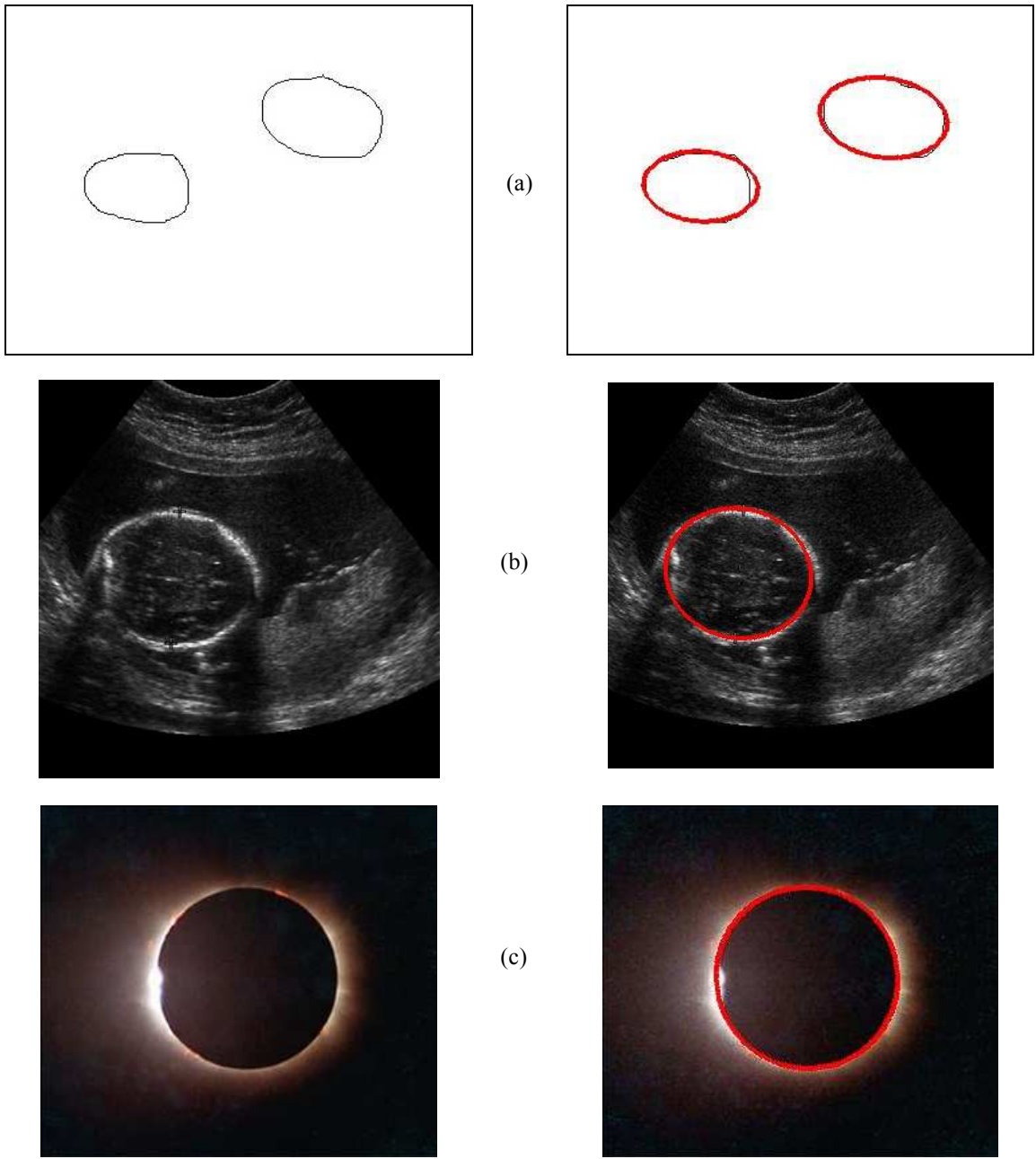

**Fig. 10** CAB Approximating ellipsoidal shapes and arc detections

*5.4. Performance comparison*

In order to enhance the performance analysis, the proposed approach is compared to the GA-based algorithm [14] and the RHT [10] method over a common image set.

The GA-based algorithm follows the proposal of Yao et al. [14], which considers two different subpopulations (each one of 50 individuals), a crossover probability of 0.55, the mutation probability being 0.10 and the number of elite individuals chosen as 5. The roulette wheel selection and the 1-point crossover operator are also applied. The parameter setup and the fitness function follow the configuration suggested in [14]. Likewise, the RHT method has been implemented as it is described in [10].

Images rarely contain perfectly-shaped ellipses. In order to test the accuracy for a single-ellipse, the detection is challenged by a ground-truth ellipse, which is determined manually from the original edge-map. The parameters $(x_0^{true}, y_0^{true}, r_{max}^{true}, r_{min}^{true}, \theta_{true})$ of the ground-truth ellipse are computed considering the best fitted ellipse that a human observer can identify through a drawing software (MSPaint©, CorelDraw©, etc). If the parameters of the detected ellipse are defined as $x_0^D, y_0^D, r_{max}^D, r_{min}^D$ and $\theta_D$, then an error score (Es) is defined as follows:

$$Es = P_1 \left( \left| x_0^{true} - x_0^D \right| + \left| y_0^{true} - y_0^D \right| \right) + P_2 \left( \frac{\left| r_{max}^{true} - r_{max}^D \right| + \left| r_{min}^{true} - r_{min}^D \right|}{2} \right) + P_3 \left| \theta_{true} - \theta_D \right| \quad (19)$$

The central point difference $\left( \left| x_0^{true} - x_0^D \right| + \left| y_0^{true} - y_0^D \right| \right)$ represents the center shift for the detected ellipse as it is compared to the ground-truth ellipse. The averaged radio mismatch $\left( \left( \left| r_{max}^{true} - r_{max}^D \right| + \left| r_{min}^{true} - r_{min}^D \right| \right) / 2 \right)$ accounts for the difference between their radii. Finally, the angle error $\left( \left| \theta_{true} - \theta_D \right| \right)$ corresponds to the orientation difference between the ground-truth and detected ellipses. $P_1, P_2$ and $P_3$ represent three weighting parameters, which are applied to the central point difference, to the radius mismatch and to the angle error in order to impose a relative importance in the final error Es. In this study, they are chosen as $P_1 = 0.05, P_2 = 0.1$ and $P_3 = 0.2$. This particular choice ensures that the angle error would be strongly weighted in comparison to the previous terms which account for the difference in the central ellipse positions and the averaged radii differences of manually detected and machine-detected ellipses, respectively. In order to use an error metric for multiple-ellipse detection, the averaged Es resulting from each ellipse in the image, is considered. The multiple error (ME) is thus calculated as follows:

$$ME = \left( \frac{1}{NC} \right) \cdot \sum_{R=1}^{NC} Es_R \quad (20)$$

where *NC* represents the number of ellipses actually present the image. In case of ME being less than 1, the algorithm is considered successful; otherwise it is said to have failed in the detection of the ellipse set. Notice that for $P_1 = 0.05, P_2 = 0.1$ and $P_3 = 0.2$, an ME<1 is generated accounting for a maximal tolerated average difference for radius (10 pixels long) and a mismatch limit of 5 degrees whereas the maximum average mismatch for the centre location can be up to 20 pixels. In general, the Success Rate (SR) can thus be defined as the percentage of achieving success after a certain number of trials.

Fig. 11 and 12 show six images (three synthetic and three natural) that have been used to compare the performance of the GA-based algorithm [14], the RHT method [10] and the proposed approach. The performance is analyzed by considering 35 different executions for each algorithm over six images. Table 2 presents the averaged execution time, the success rate (SR) in percentage and the averaged multiple error (ME). The best entries are bold-cased in Table 2. Closer inspection reveals that the proposed method is able to achieve the highest success rate with the smallest error and still requires less computational time for most cases. Fig. 11 and 12 also exhibit the resulting images after applying the correspondent detector over each image. Such results present the median cases obtained throughout 35 runs.

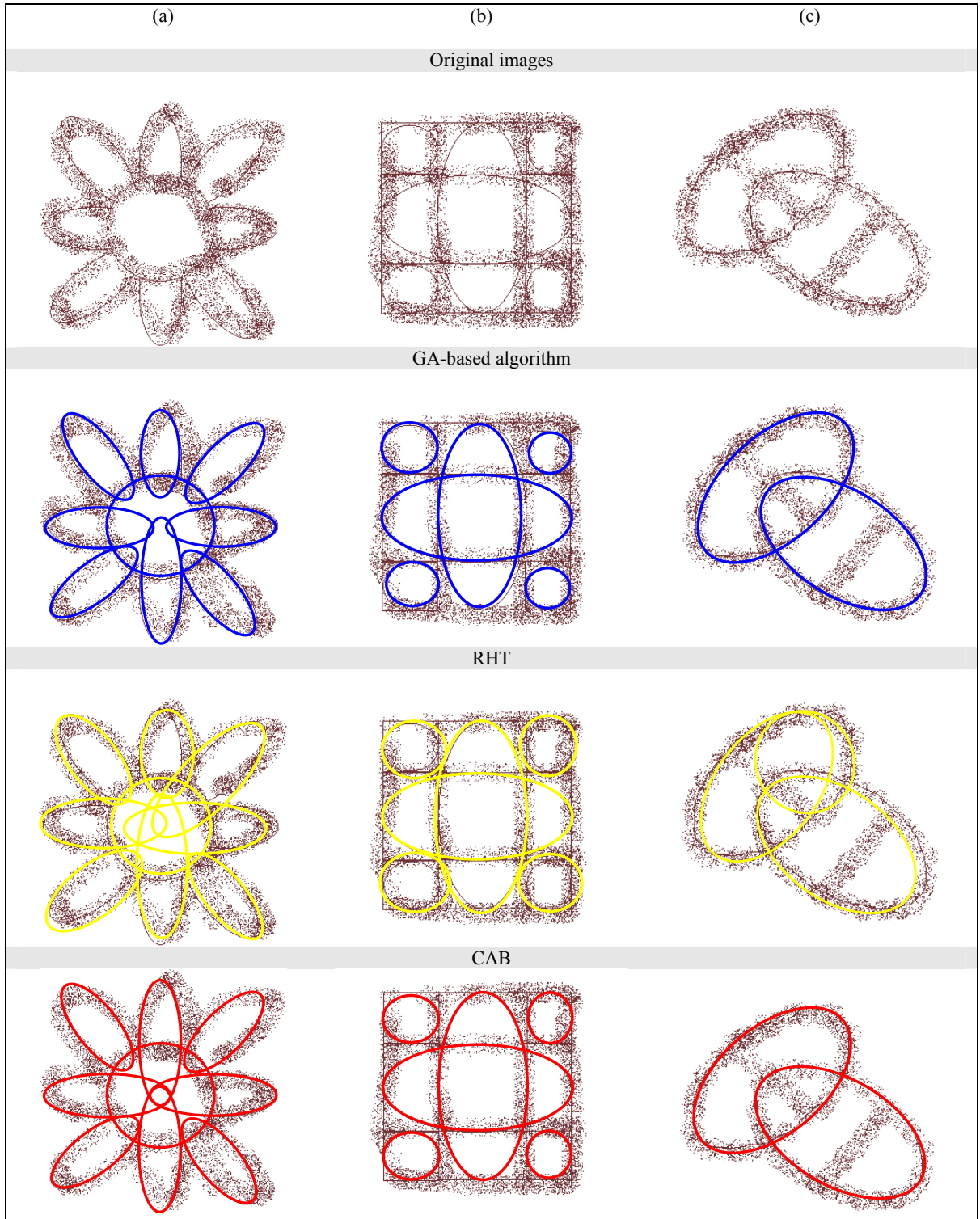

**Fig. 11.** Synthetic images and their detected ellipses for: GA-based algorithm, the RHT method and the proposed CAB algorithm.

 

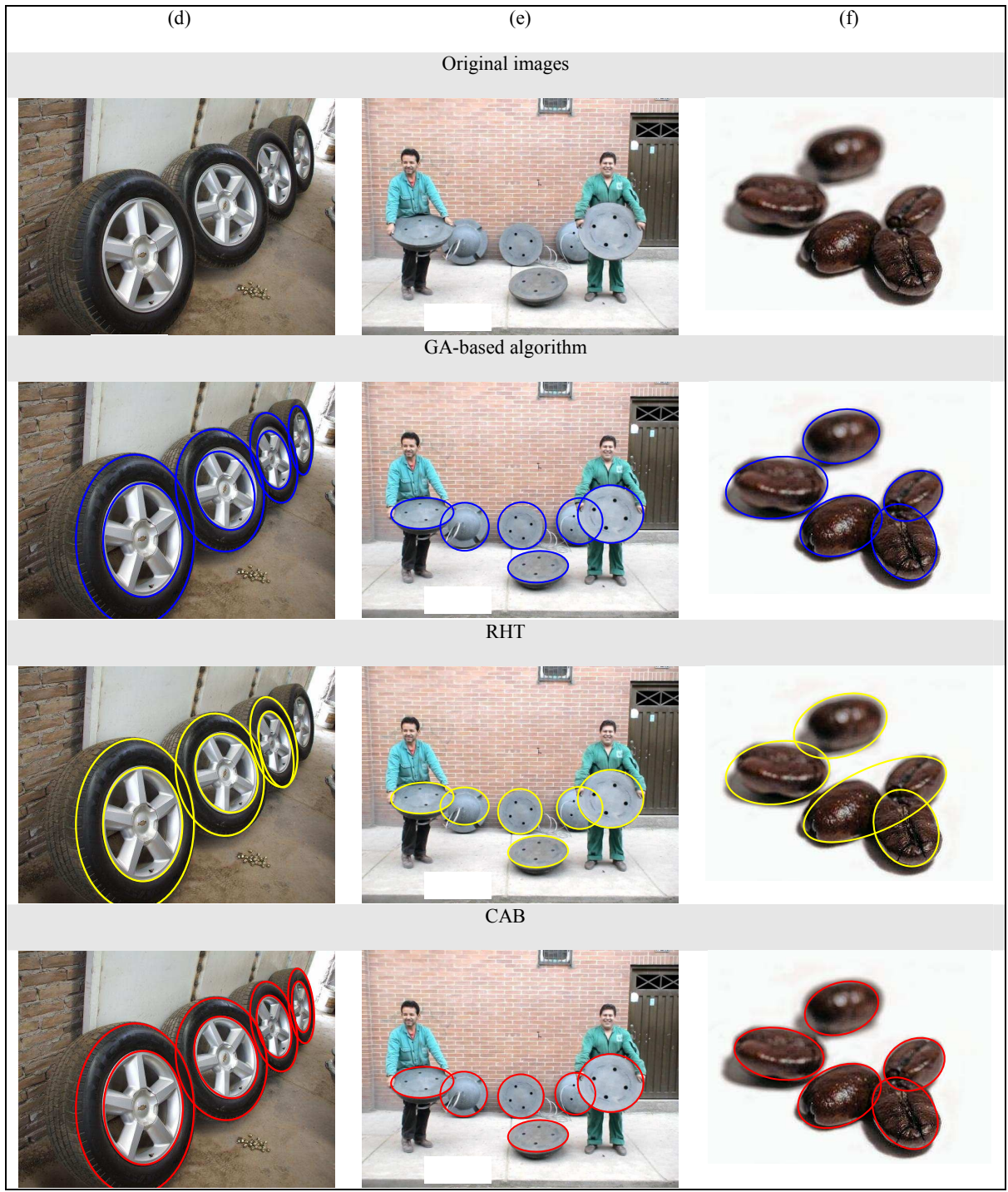

**Fig. 12.** Real-life images and their detected ellipses for: GA-based algorithm, the RHT method and the proposed CAB algorithm.

| Image | Averaged execution time ± Standard deviation (s) | | | Success rate (SR) (%) | | | Averaged ME ± Standard deviation | | |
|---|---|---|---|---|---|---|---|---|---|
| | GA | RHT | CAB | GA | RHT | CAB | GA | RHT | CAB |
| (a) | 4.32±(0.81) | 2.82±(0.77) | **0.51±(0.31)** | 94 | 90 | **100** | 0.78±(0.023) | 0.98±(0.076) | **0.31±(0.052)** |
| (b) | 6.83±(1.23) | 3.75±(0.92) | **0.62±(0.63)** | 97 | 88 | **100** | 0.85±(0.053) | 0.93±(0.087) | **0.19±(0.027)** |
| (c) | 5.28±(0.11) | 3.18±(0.89) | **0.57±(0.68)** | 100 | 91 | **100** | 0.89±(0.096) | 0.92±(0.051) | **0.26±(0.018)** |
| (d) | 7.08±(1.27) | 3.82±(0.67) | **0.60±(0.42)** | 79 | 70 | **100** | 0.84±(0.039) | 0.89±(0.092) | **0.19±(0.061)** |
| (e) | 6.83±(1.77) | 3.55±(0.54) | **0.57±(0.36)** | 100 | 89 | **100** | 0.92±(0.058) | 0.98±(0.023) | **0.29±(0.061)** |
| (f) | 6.77±(1.23) | 3.88±(0.78) | **0.54±(0.51)** | 93 | 90 | **100** | 0.80±(0.092) | 0.88±(0.062) | **0.28±(0.071)** |

**Table 2.** The averaged execution-time, success rate and the averaged multiple error for the GA-based algorithm, the RHT method and the proposed CAB algorithm, considering the six test images shown in Fig. 9

In order to statistically analyze the results in Table 2, a non-parametric significance proof known as the Wilcoxon's rank test [35]-[37] has been conducted. Such proof allows assessing result differences among two related methods. The analysis is performed considering a 5% significance level over multiple error (ME) data. Table 3 reports the *p*-values produced by Wilcoxon's test for a pair-wise comparison of the multiple error (ME), considering two groups gathered as CAB vs. GA and CAB vs. RHT. As a null hypothesis, it is assumed that there is no difference between the values of the two algorithms. The alternative hypothesis considers an existent difference between the values of both approaches. All *p*-values reported in the Table 3 are less than 0.05 (5% significance level) which is a strong evidence against the null hypothesis, indicating that the best CAB mean values for the performance are statistically significant which has not occurred by chance.

| Image | *p*-Value | |
|---|---|---|
| | CAB vs. GA | CAB vs. RHT |
| Synthetic images | | |
| (a) | 1.7401e-004 | 1.7245e-004 |
| (b) | 1.4048e-004 | 1.6287e-004 |
| (c) | 1.8261e-004 | 1.4287e-004 |
| Natural Images | | |
| (a) | 1.4397e-004 | 1.0341e-004 |
| (b) | 1.5481e-004 | 1.6497e-004 |
| (c) | 1.7201e-004 | 1.4201e-004 |

**Table 3.** *p*-values produced by Wilcoxon's test comparing CAB to GA and RHT over the averaged ME from Table 2.

## 6. Conclusions

This paper discusses a novel and effective technique for extracting multiple ellipses from an image that considers the overall detection process as a multi-modal optimization problem. In the detection, the approach employs an evolutionary algorithm based on the way animals behave collectively. In such algorithm, searcher agents emulate a group of animals which interact to each other using simple biological rules that are modeled as evolutionary operators. Such operators are applied to each agent considering that the complete group has a memory to store the optimal solutions (ellipses) seen so-far by applying a competition principle. The detector uses a combination of five edge points as parameters to determine ellipse candidates (possible solutions). A matching function determines if such ellipse candidates are actually present in the image. Guided by the values of such matching functions, the set of encoded candidate ellipses are evolved through the evolutionary algorithm so that the best candidates can be fitted into actual ellipses within the image. Just after the optimization process ends, an analysis over the embedded memory is executed in order to find the best obtained solution (the best ellipse) and significant local minima (remaining ellipses). The overall approach generates a fast sub-pixel detector which can effectively identify multiple ellipses in real images despite ellipsoidal objects exhibiting a significant occluded or distorted portion.

Classical Hough transform methods for ellipse detection use five edge points to cast a vote for the potential elliptical shape in the parameter space. However, they would require a huge amount of memory and longer computational time to obtain a sub-pixel resolution. Moreover, HT-based methods rarely find a precise parameter set for a given ellipse in the image [38]. In our approach, the detected ellipses hold a sub-pixel accuracy inherited directly from the ellipse equation and the MEA method.

In order to test the ellipse detection performance, both its speed and accuracy have been compared. Score functions are defined by Eqs. 19 and 20 in order to measure accuracy and effectively evaluate the mismatch between manually detected and machine-detected ellipses. We have demonstrated that the CAB method outperforms both the GA (as described in [14]) and the RHT (as described in [10]) within a statistically significant framework (Wilcoxon test).